\pdfoutput=1

\documentclass[11pt]{article}

\usepackage{EMNLP2023}

\usepackage{times}
\usepackage{latexsym}

\usepackage[T1]{fontenc}

\usepackage[utf8]{inputenc}

\usepackage{microtype}

\usepackage{inconsolata}

\usepackage{amsmath, tabularx, longtable, multirow}
\DeclareMathOperator{\Score}{score}

\usepackage{graphicx}
\usepackage{adjustbox, booktabs}
\usepackage{rotating}

\title{On Using Distribution-Based Compositionality Assessment to Evaluate Compositional Generalisation in Machine Translation}

\author{Anssi Moisio$^{1,2}$, Mathias Creutz$^2$, \and Mikko Kurimo$^1$ \\
        $^1$Department of Information and Communications Engineering, Aalto University, Finland \\
        $^2$Department of Digital Humanities, University of Helsinki, Finland \\
        \texttt{anssi.moisio@aalto.fi}, \texttt{mathias.creutz@helsinki.fi}, \texttt{mikko.kurimo@aalto.fi}}

\begin{document}
\maketitle
\begin{abstract}
Compositional generalisation (CG), in NLP and in machine learning more generally, has been assessed mostly using artificial datasets. It is important to develop benchmarks to assess CG also in real-world natural language tasks in order to understand the abilities and limitations of systems deployed in the wild. To this end, our GenBench Collaborative Benchmarking Task submission utilises the \emph{distribution-based compositionality assessment} (DBCA) framework to split the Europarl translation corpus into a training and a test set in such a way that the test set requires compositional generalisation capacity. Specifically, the training and test sets have divergent distributions of dependency relations, testing NMT systems' capability of translating dependencies that they have not been trained on. This is a fully-automated procedure to create natural language compositionality benchmarks, making it simple and inexpensive to apply it further to other datasets and languages. The code and data for the experiments is available at \url{https://github.com/aalto-speech/dbca}.
\end{abstract}

\section{Introduction}

An often-used definition, by \citet{partee1995lexical}, of the concept of \emph{compositionality} of language is that the ``meaning of a whole is a function of the meanings of the parts and of the way they are syntactically combined''.
A more stringent definition adds that the composition of meaning is \emph{systematic}: each part's meaning is the same in all the different sequences it appears in, the syntactical rules work the same way for different parts, and the same function determines meaning for different wholes \citep{fodor1988connectionism,pavlick2022slides,pavlick2023symbols}. Compositionality enables generalising to new meanings by combining familiar parts, and to understand each other language users need to employ common systematic rules.

A number of benchmarks have been developed to assess systematic generalisation from different perspectives and in different NLP tasks. Many of these consist of artificial data, such as the popular SCAN \citep{lake2018generalization} and COGS \citep{kim-linzen-2020-cogs} benchmarks. These artificial datasets are typically constructed to be highly systematic, to include straightforward syntactical rules. Natural languages, however, have varied irregularities, idiomatic expressions, and other exceptions to the rules, which make composition of meaning much more complicated than in the case of the highly-regular artificial datasets. It's therefore important to assess whether NLP systems are able to generalise systematically also in the case of natural language, where systematic rules are obscured by exceptions \citep{dankers2022paradox}.

Works that aim to assess systematic generalisation in natural language tasks often still synthesise some part of the dataset in order to create test examples where systematic generalisation is needed (for example \citet{li-etal-2021-compositional} and \citet{dankers2022paradox}). This enables precise testing of specific systematic rules, but comprehensive test suites that would assess the use of, for example, numerous syntactical rules can be arduous to synthesise. To sidestep the need to synthesise examples, a natural language dataset can be partitioned into training and test sets in such a way that the test set includes examples whose processing requires some systematic generalisation ability. A framework for partitioning data in this way was developed by \citet{keysers2019measuring}, called \emph{distribution-based compositionality assessment}, or DBCA for short. The main idea of DBCA is to control the distributions of \emph{atoms} (primitive elements) and \emph{compounds} (combinations of the atoms) to get approximately the same atom distributions but divergent compound distributions in the training and test sets.

\begin{table*}[htb]
\footnotesize
\centering
\begin{tabular}{p{5cm}|p{4cm}|p{4.4cm}}
 \toprule
 Sentence & Atoms & Compounds \\ \midrule
    ``\emph{Our vigilance is not partisan.}''
                & \texttt{nsubj, poss, our, vigilance, partisan}
                & \texttt{(vigilance, poss, our), (partisan, nsubj, vigilance)} \\ \midrule
    ``\emph{We shall now hear Mr Wurtz speaking against this request.}''
                & \texttt{hear, aux, shall, speak, nsubj, wurtz, hear, ccomp, speak}
                & \texttt{(hear, aux, shall), (speak, nsubj, wurtz), (hear, ccomp, speak)} \\ \midrule
    ``\emph{This seems to me to be a workable solution.}''
                & \texttt{solution, amod, workable, seem, xcomp, solution}
                & \texttt{(solution, amod, workable), (seem, xcomp, solution)} \\
\bottomrule
\end{tabular}
\caption{Examples of what we call ``atoms'' and ``compounds''. Atoms are the lemmas and dependency relations, and compounds the three-element tuples of the head lemma, the relation, and the dependant lemma.} \label{tab:examples}
\end{table*}

We utilise the DBCA framework in our GenBench Collaborative Benchmarking Task submission,  which consists of train-test splits of the Europarl parallel corpus with divergent distributions of dependency relations. These data splits can be used to assess the ability of NMT systems to translate novel dependency relations. In the terminology of the DBCA framework, we define the atoms as lemmas and dependency relations, and the compounds as the three-element tuples of the head lemma, the dependant lemma, and the relation between them (see Table~\ref{tab:examples} for examples). This method to create compositional generalisation benchmarks does not require manual test suite construction, making it easy to extend it to other datasets and other languages.

\section{Related work}

\subsection{Compositional generalisation in machine translation}

Compositional generalisation has been assessed in machine translation in a few works in recent years.
\citet{raunak2019compositionality} partitioned a dataset into short training sentences and longer test sentences in order to assess generalisation from short sentences to longer ones, a subtype of compositional generalisation sometimes called \emph{productivity} \citep{hupkes2020compositionality}.
\citet{li-etal-2021-compositional} synthesised sentences for the test set with novel constituents, such as noun and verb phrases to create the CoGnition benchmark.
\citet{dankers2022paradox} assessed three aspects of compositionality in NMT, which they called systematicity, the ability to combine familiar parts into novel combinations; substitutivity, the consistency of translations when a word is replaced with its synonym; and over-generalisation, the tendency to follow a compositional rule even when the case is actually an exception to the rule.

Perhaps the most similar benchmark to ours is ReaCT by \citet{zheng-lapata-2023-real}. In ReaCT, the IWSLT 2014 German-English corpus is used as the training set, and a test set is created by selecting sentences that have a high \emph{compositionality degree} from the WMT 2014 corpus.
The compositionality degree of a test set sentence is defined as the number of training set n-grams needed to create the sentence, divided by the length of the sentence. The reasoning is that a test set sentence that includes long n-grams from the training set can be composed of fewer n-grams, having a low compositionality degree, whereas if we need to back off to shorter n-grams to create a sentence, the compositionality degree is high. This train-test data split has similar general characteristics as our data splits: it is completely natural data (no synthetic sentences) partitioned so that the test set has novel combinations of familiar primitives. In contrast to our work, the primitives in this scheme can be sequences of multiple words, whereas we assess the ability to translate novel combinations of just two words and their dependency relation. We also pay attention to the relative frequency distributions of the primitives and the combinations, following the DBCA framework, whereas in ReaCT the compositionality score is a function of the unique n-gram types.

In addition to word-based compositionality, \emph{morphological} compositional generalisation in translation has been assessed by \citet{meyer-buys-2023-subword} and \citet{moisio-etal-2023-evaluating}. In these works, the atoms are defined as the morphemes (or surface-level morphs), and compounds as the inflected word forms that consist of multiple morphemes. The test set therefore includes novel combinations of familiar morphemes, assessing NMT systems' capacity for morphological generalisation.  

\subsection{Other related work}

\begin{figure*}[htb]
\centering
\includegraphics[scale=0.28]{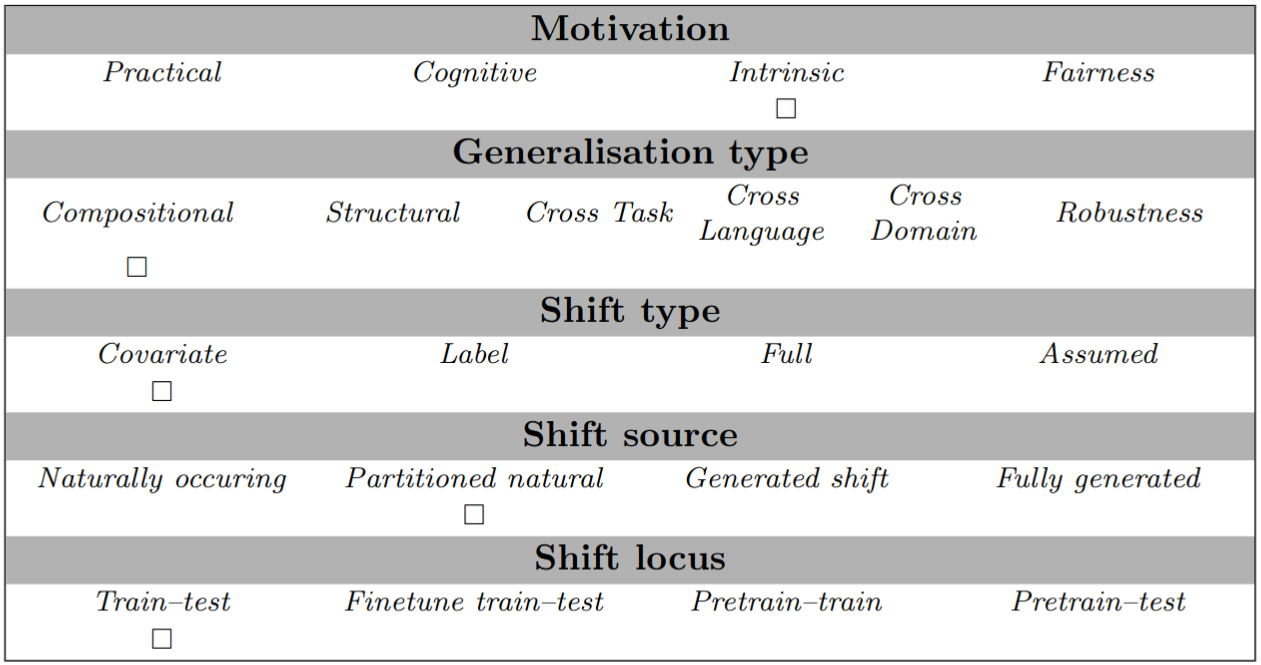}
\caption{The categorisation of our benchmark in the taxonomy by \citet{hupkes2022taxonomy}.}
\label{fig:evalcard} 
\end{figure*}

Aside from NMT tasks, \citet{shaw2021compositional} utilised the DBCA framework to assess compositional generalisation in a natural language semantic parsing task. \citet{sogaard2021we} provide a more general discussion and review of non-random training-test data splitting.

Besides creating artificial train-test splits, another option to test systematic generalisation in NLP systems, without the need for manual test suite design, is to leverage the fact that systematicity can be seen as an inherent symmetry in the data \citep{manino-etal-2022-systematicity}, which has also been utilised to generate new training examples \citep{akyurek-andreas-2023-lexsym}.

A study not on compositional generalisation, but in other ways related to our work, is that by \citet{mccoy-etal-2023-much}, who evaluated the degree of novelty of the text generated by language models, using both n-grams and dependency relations. They found that language-model-generated text tends to be less novel than the baseline of human-generated text in local structure (small n-grams), but more novel than the human baseline in more global structure (large n-grams). This finding provides a good backdrop for our research question of how NMT models handle a test set that includes novel \emph{local} structure, such as dependency relations.

\subsection{The taxonomy}
The eval card in Figure~\ref{fig:evalcard} shows how our benchmark can be categorised in the taxonomy of
\citet{hupkes2022taxonomy}. The motivation is primarily intrinsic: it is important to assess if translation models learn the systematic rules that characterise natural language, in order to get some understanding how the models work. Another motivation is practical; testing compositional generalisation is important for the practical reason of knowing how robustly the models generalise to novel dependency relations. The type of the generalisation is compositional, and the shift type is covariate, since the input data distribution changes but the task remains otherwise the same. Shift source is partitioned natural data, since we do not use any artificial data, but the train-test split is artificial. Lastly, the shift locus in our experiments is train-test, but the method and benchmark could also possibly be used as a finetune train-test benchmark, by finetuning a pretrained model on the training set.

\section{Europarl data splits} \label{sec:splits} 

\subsection{Data partitioning process}

We use the Europarl corpus \citep{koehn-2005-europarl} of transcribed European parliament proceedings, with the multilingual sentence alignments from the OPUS corpus \citep{TIEDEMANN2012opus}. We chose the Europarl corpus because of the good quality of the translations and because it includes parallel sentences for multiple languages. For our benchmark submission, we select English as the source language, and as the target languages we use four languages that represent different (branches of) language families: German, French, Greek, and Finnish.

As pre-processing, duplicate sentences are removed and maximum sentence length is restricted to 30 words before tokenisation. We take a random subsample of 300k sentences from which we extract the data splits. This relatively small size was chosen for convenient use as well as to allow comparison with previously published similar benchmarks: CoGnition \citet{li-etal-2021-compositional} and ReaCT \citet{zheng-lapata-2023-real}, which are similar in size.

The dependency parsing for the English source corpus is done using the LAL-parser \citep{mrini-etal-2020-rethinking}. To calculate the divergences in data splitting, we only consider the English side. Therefore, the benchmark primarily assesses the encoder's capacity to represent novel syntactic relations. However, presumably a high compound divergence of the source side sentences means that the target side sentences also include an increased number of novel syntactical, or possibly morphological, structures, assessing at the same time the decoder's capacity to generate these. We define the atoms as the lemmas and dependency relations and compounds as the three-element-tuples of the dependant lemma, the head lemma and their relation. To make the number of atoms manageable, we exclude from the distribution calculations lemmas that appear either very frequently (the 200 most frequent lemmas, which each appear from 387k times (most frequent word ``the'') to 3576 times (200th most frequent word ``then'')), or fewer than 10 times in total in the corpus. After this filtering, about 8000 lemmas remain in the atom set, which includes also the dependency relation tags.

Following \citet{keysers2019measuring}, we calculate a weight for each compound so that those sub-compounds that appear predominantly only in one super-compound get lower weight than those appearing in many different super-compounds. In our case, this means that if a (dependant lemma, relation type) pair occurs for example 8/10 times with just one head lemma, this pair gets a score of $1-(8/10)=0.2$. The idea is somewhat similar to that behind the Kneser-Ney smoothing \citep{kneser1995improved}, where the number of different bigrams a word appears in correlates with the unigram backoff probability. We filter out compounds that get a weight less than 0.5, after which there are about 8000 atom types and 400k compound types whose distributions are controlled in the data splits.

Atom and compound divergences are calculated similarly to \citet{keysers2019measuring}: divergence $\mathcal{D}$ between distributions $P$ and $Q$ is calculated using the Chernoff coefficient $C_\alpha(P \Vert Q) = \sum_{k} p_k^\alpha \, q_k^{1-\alpha} \in [0, 1]$ \citep{chung1989measures}, with $\alpha=0.5$ for the atom divergence and $\alpha=0.1$ for the compound divergence. \citet{keysers2019measuring} notes about the $\alpha$ values that $\alpha=0.5$ for atom divergence ``reflects the desire of making the atom distributions in train and test as similar as possible'', and $\alpha=0.1$ for compound divergence ``reflects the intuition that it is more important whether a certain compound occurs in P (train) than whether the probabilities in P (train) and Q (test) match exactly''. The divergence is the complement of the Chernoff coefficient, since the Chernoff coefficient measures similarity between two vectors. The atom and compound divergences for training set $V$ and test set $W$ are:
\begin{align*}
    \mathcal{D}_A(V \Vert W) &= 1\,-\, C_{0.5}(\mathcal{F}_A(V) \, \Vert \, \mathcal{F}_A(W)) \\
    \mathcal{D}_C(V \Vert W) &= 1\,-\, C_{0.1}(\mathcal{F}_C(V) \, \Vert \, \mathcal{F}_C(W)).
\end{align*}
where $\mathcal{F}_A$ is the atom distribution and $\mathcal{F}_C$ is the compound distribution of a data set.

Splitting the data is done using a greedy algorithm similar to that by \citet{moisio-etal-2023-evaluating}. This algorithm places one sentence at each iteration into either the training or test set, such that the atom and compound divergences are as close to the respective desired values as possible. Specifically, at each iteration we try to maximise a score that is the negated linear combination of the differences between the desired and actual divergence values:
\begin{align*}
    \Score(V, W) = - | c - \mathcal{D}_C(V \Vert W) | - \mathcal{D}_A(V \Vert W),
\end{align*}
where $c$ is the desired compound divergence, and the desired atom divergence is 0 (minimum).

\subsection{Comparison to random splits and previous benchmarks}

\begin{table*}[htb]
\centering
\begin{tabular}{l|cc|cc|c|cc}
\toprule
& \multicolumn{2}{c|}{\# sentences} & \multicolumn{2}{c|}{\# words } & \# unique lemmas & $\mathcal{D}_A $ & $\mathcal{D}_C$ \\ \midrule
& train & test & train & test &  & &  \\ \midrule
Europarl random split 3k                        & 200k & 3k  & 3.6M  & 54k  & 29k  & 0.28  & 0.60 \\
Europarl random split 10k                       & 200k & 10k & 3.6M  & 180k & 30k  & 0.18  & 0.55 \\
Europarl random split 30k                       & 200k & 30k & 3.6M  & 540k & 31k  & 0.13  & 0.52 \\ \midrule
CoGnition \citep{li-etal-2021-compositional}    & 196k & 10k & 1.9M  & 96k  & 1.7k & 0.13  & 0.47 \\
ReaCT \citep{zheng-lapata-2023-real}            & 160k & 3k  & 3.3M  & 45k  & 53k  & 0.32  & 0.90 \\ \midrule
Europarl min$\mathcal{D}_C$ split \#1           & 203k & 37k & 3.9M  & 650k & 34k  & 0.01  & 0.10 \\
Europarl min$\mathcal{D}_C$ split \#2           & 194k & 36k & 3.7M  & 625k & 34k  & 0.01  & 0.10 \\
Europarl min$\mathcal{D}_C$ split \#3           & 195k & 35k & 3.7M  & 625k & 34k  & 0.01  & 0.10 \\
Europarl max$\mathcal{D}_C$ split \#1           & 197k & 23k & 3.8M  & 390k & 34k  & 0.001 & 1.0 \\
Europarl max$\mathcal{D}_C$ split \#2           & 198k & 22k & 3.8M  & 390k & 34k  & 0.002 & 1.0 \\
Europarl max$\mathcal{D}_C$ split \#3           & 198k & 22k & 3.8M  & 390k & 34k  & 0.001 & 1.0 \\
\bottomrule
\end{tabular}
\caption{Comparison of the Europarl splits to other translation benchmarks that aim at assessing compositional generalisation. The number of unique lemmas includes both training and test set lemmas.} \label{tab:comparison}
\end{table*}

Table~\ref{tab:comparison} lists the sizes, and atom and compound divergences ($\mathcal{D}_A$ and $\mathcal{D}_C$) for the Europarl data splits, as well as for random splits, and for the CoGnition and ReaCT benchmarks for comparison. All the divergences are calculated after similar filtering of the atoms and compounds as described for the Europarl splits above. From the divergences of the random train-test splits we can notice that the size of the test set correlates inversely with both atom and compound distributions; when the training and test sets are closer in size, the distributions are also naturally closer to each other.

From the table, we can also see that CoGnition includes relatively short sentences and, importantly, a relatively small number of unique lemmas. Creating CoGnition, \citet{li-etal-2021-compositional} removed some of the complexity of natural language, such as polysemous words, and deliberately kept the vocabulary small and excluded all low-frequency words. This follows the design choices made by \citet{keysers2019measuring} of aiming to have only few meaningful atoms from which a large number of compounds can be created, which was motivated by practical concerns: this way it is easier to have a large range of compound divergences while keeping the atom divergences same. However, this contrasts with the distribution of primitives in natural language, an issue we discuss more in Section~\ref{sec:discussion}. ReaCT, on the other hand, has similarly sized vocabulary as natural language data; in fact the vocabulary is significantly larger than the Europarl random sample vocabulary, possibly because the WMT and IWSLT corpora are lexically more diverse than Europarl.

Table~\ref{tab:comparison} also lists, for reference, the atom and compound divergences, calculated for the dependency relations as described in Section~\ref{sec:splits}, for CoGnition and ReaCT, even though neither of these data splits are designed to minimise or maximise these values. In both of these data sets, the test set is designed to contain novel combinations of familiar parts, as is our test sets, but in these works the parts are normally sequences of multiple words. In spite of this, the ReaCT data split has a relatively high dependency relation compound divergence.

The last rows of Table~\ref{tab:comparison} show the sizes and divergences of the minimum- and maximum-compound-divergence Europarl data splits. The atom divergences are significantly lower for these splits than they are for the random splits. As noted above, the low atom divergence follows the principles of the DBCA framework: the compositional generalisation test set should be difficult not because of novel primitive elements (in our case mostly lemmas) but because of \emph{novel combinations} of the known elements. The maximum-compound-divergence splits get a $\mathcal{D}_C$ of exactly 1, but here we note that, as described in Section~\ref{sec:splits}, the most frequent and most infrequent lemmas are left out from the divergence calculations, which means that only a subset of the dependency relation compounds are novel in the test set, even though $\mathcal{D}_C=1.0$.

\section{Experiments}

\subsection{Transformer NMT system results} 

\begin{figure*}[htb] 
\hspace*{-0.3cm}  
\centering
\includegraphics[scale=0.51]{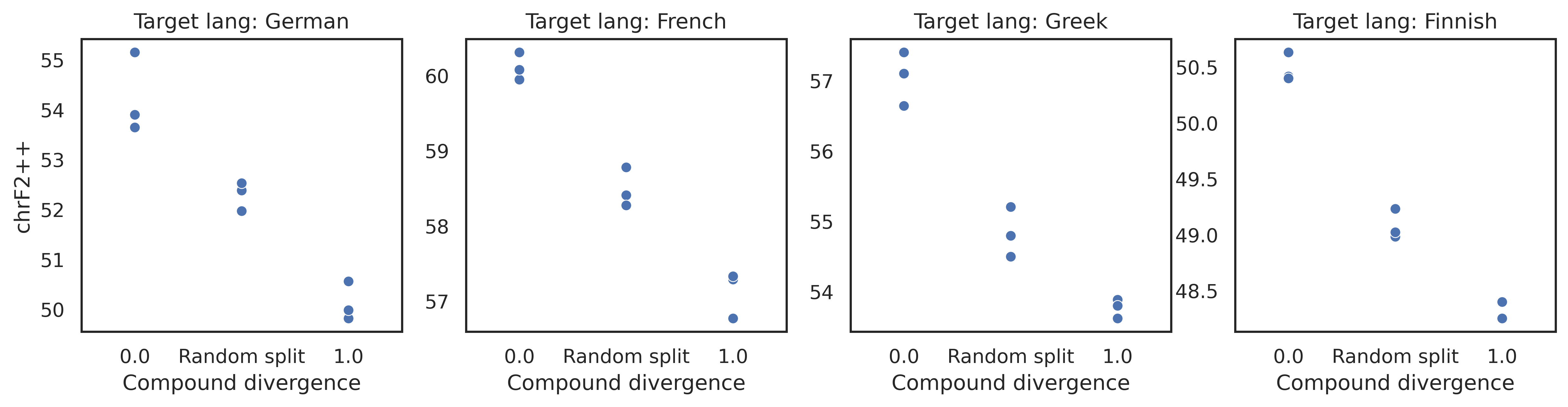}
\caption{The chrF2++ scores for the Transformer NMT systems trained on the minimum and maximum compound divergence splits. Source language is English and target languages are German, French, Greek, and Finnish. The middle scores are for the random data splits, which happen to have a compound divergence of about 0.5. Each data split setup (min$\mathcal{D}_C$, max$\mathcal{D}_C$, and random) is run 3 times with different random seeds, creating 9 different data splits for which NMT models are trained.}
\label{fig:nmt_scatter} 
\end{figure*}

We train Transformer \citep{vaswani2017attention} models on the Europarl data splits using the OpenNMT python library \citep{klein-etal-2017-opennmt} and the same hyperparameters as in the example OpenNMT-py Transformer configuration\footnote{\url{https://github.com/OpenNMT/OpenNMT-py/blob/9d617b8b/config/config-transformer-base-1GPU.yml}}, including the standard 6 transformer layers with 8 heads, a hidden layer size of 512 and feed-forward layer size of 2048. The configuration includes around 60M parameters. We create a BPE \citep{sennrich2016neural} vocabulary for each language with 10k token types. We didn't tune any of the hyperparameters for these datasets, and didn't use a validation set. We trained the models for 6000 training steps, evaluated the test set translations at intervals of 1000 steps, and report the best of these 6 results for each model. For more details on training the models, see the Github repository linked on the first page.

Figure~\ref{fig:nmt_scatter} displays the chrF2++ \citep{popovic2017chrf} scores for the minimum- and maximum-compound-divergence splits and the four target languages: German, French, Greek, and Finnish. We run the data split algorithm three times using different random seeds. Between the min$\mathcal{D}_C$ and max$\mathcal{D}_C$ split results are the results for the random splits, of which there are also 3 random runs (with 200k sentences in training and 30k sentences in test set). As shown in Table~\ref{tab:comparison} the random splits with 30k-sentence test set happen to have a compound divergence of about 0.5.

Figure~\ref{fig:nmt_scatter} shows a modest but statistically significant and consistent decrease in performance from the random data split to the max$\mathcal{D}_C$ split, for all four target languages. This is expected, as the max$\mathcal{D}_C$ split includes more novel dependency relations than the random split. However, as shown in Table~\ref{tab:comparison}, the max$\mathcal{D}_C$ split has a significantly lower \emph{atom} divergence than the random split, as this is deliberately minimised in the artificial data splits, which follows the principle of the DBCA framework of having the same atom distribution in training and test sets. This means that the max$\mathcal{D}_C$ split should be easier than the random split in this regard, but it still gets worse results because of the high compound divergence. 

The min$\mathcal{D}_C$ split, on the other hand, has a similar atom divergence as the max$\mathcal{D}_C$ split, so comparison between these two results is in that sense fairer. There is a larger difference in the results between these two data splits; depending on the target language the chrF2++ drops from about 4\% to 8\%. Since we use relatively large test corpora (from about 20k to 40k sentences), even small differences in chrF2++ are statistically significant.

\subsection{Generalisation score}

To assess whether one NMT system is more capable in (this dependency-relation-related type of) compositional generalisation than some other system, one option is simply to compare their translation performances on the max$\mathcal{D}_C$ split. However, to get a sense of the generalisation capacity as a part of the system's translation capacity in general, it may be more meaningful to assess how the performance deteriorates between the min$\mathcal{D}_C$ and max$\mathcal{D}_C$ splits. To get a \emph{generalisation score}, we propose to take the ratio of the results on these two data splits. This way the generalisation scores, using the average of the three chrF2++ scores for each experiment setup listed in Figure~\ref{fig:nmt_scatter}, are: $50.12/54.23=0.92$ for the German-target Transformer system, $57.13/60.11=0.95$ for French, $53.77/57.05=0.94$ for Greek, and $48.30/50.48=0.96$ for Finnish. Since this a relative score, the absolute chrF2++ results should be reported in addition to this generalisation score.

\section{Discussion: handling the long tail} \label{sec:discussion}

The core of each natural language is a set of words (a lexicon), and a set of grammar rules that define how to combine the words into meaningful sequences. The lexicon is divided into content words, which possess semantic content, and function words, which denote the grammatical relationships between content words. The function words belong to closed classes, for example prepositions, that normally do not accept new words, while content words belong to open classes; new nouns, for example, are coined regularly. The long tail of the Zipfian distribution of word frequencies consists of content words while the grammar-enforcing function words are mostly in the head of the distribution.

Recent studies suggest that the distinction between content and function words as well as the Zipfian distribution are important for (compositional) generalisation to arise: \citet{steinert2020toward} provides empirical evidence that function words enable robust forms of non-trivial compositional communication, and \citet{chan2022data} found that the Zipfian distribution helps language models strike a balance between memorisation and (in context) generalisation. As shown also by \citet{feldman2020does}, memorising some of the long tail is not in conflict with generalisation; on the contrary, it \emph{enables} optimal generalisation.

At the same time, some studies have shown that neural NLP system have some difficulties with handling the long tail, or at least handling the tail in a way that we as users of the models would expect and want. \citet{wei-etal-2021-frequency} and \citet{czarnowska-etal-2019-dont} found that NLP systems' performance is heavily influenced by word frequency in training. \citet{lebrun2021evaluating} compared language-model-generated text to a reference and found that the models underestimate the long tail of well-formed sequences; furthermore, this probability mass didn't go to the head of the distribution but rather the models overestimate the probability of ill-formed sequences.

As noted in previous sections, the desired distribution of atoms in the DBCA framework contrasts with the Zipfian distribution found in natural language. As \citet{keysers2019measuring} explain, they design their CFQ benchmark ``so as to have few and meaningful atoms'' which means there is no long tail of infrequent primitives. This is related to having a close-to-zero atom divergence: if the atom distribution had a long tail, it would not be easy to have the same relative atom distributions in training and test sets since at least those atoms that appear only once would make the distributions diverge. This harks back to the question of how compositional generalisation could be assessed in purely natural tasks, where every rule has an exception, and where idioms and irregularities muddle the systematicity (as discussed by \citet{dankers2022paradox}).

Our benchmark provides one answer to this question. Although we use the principles of the DBCA framework regarding distribution divergence, we don't make the corpus less natural by artificially shrinking the vocabulary size of the corpus. Instead, to make the divergence calculations manageable in practice, we leave some of the vocabulary out of the calculations. A downside of this choice is that the test sentences in the max$\mathcal{D}_C$ splits don't contain exclusively novel dependencies. The advantage is that the vocabulary is similar to that in the original real-world natural language dataset, while the test set includes an increased number of novel dependencies to test generalisation.

\section{Conclusion}

Our GenBench Collaborative Benchmarking Task submission consists of train-test splits of the  Europarl parallel corpus with divergent distributions of dependency relations. These data splits can be used to assess the ability of NMT systems to translate novel dependency relations in a purely natural language translation task. We derive the data partitioning method from the distribution-based compositionality assessment framework, which provides generalisable principles of how to assess compositionality. Our application of the DBCA framework is straightforward to extend further to new datasets and languages, and to any other NLP task where the training and test sets consist of sentences, such as paraphrase detection. The DBCA framework is useful for real-world natural language tasks too, even though it was originally designed for a more artificial data setting. It should be kept in mind, however, how the principles of the DBCA framework diverge from the reality of natural language data.

\section{Limitations}

\subsection{Applicability of the method}
Nowadays, the state-of-the-art methods in many NLP tasks are based on pretrained language models.
However, our data partitioning method, as used in this work, requires controlling the training data, as well as the test data, to have partitions with specific atom and compound divergence values. Therefore, the method is not directly applicable to a pretrain-finetune training scheme, if the pretraining dataset is not modifiable. However, there are no limitations, in principle, on having a fixed training corpus and compiling only a new test set to have specific divergence values, although the divergence values might not be so easy to minimise and maximise in this case.

\subsection{Validity of the method}

We have not conclusively assessed whether the benchmark actually tests what we assume it tests, that is, compositional generalisation ability. To rule out the most obvious potential confounding variable, we checked the sentence lengths to see if the max$\mathcal{D}_C$ test sets for some reason included longer sentences, making the test set more difficult this way. We did not find large differences: in the min$\mathcal{D}_C$ splits the average sentence lengths in train/test sets are 19.3/17.6 words, and in the max$\mathcal{D}_C$ splits 19.1/17.3 words. Although there is a difference (of unknown origin) between train and test set sentence lengths, there is no significant difference between min$\mathcal{D}_C$ and max$\mathcal{D}_C$ splits that could confound the results. From Table~\ref{tab:comparison} we can also see that the training sets are similar in size.

\subsection{Limitations of the experiments}

There are multiple levels of compositionality in language, from morphemes to words to phrases to clauses. Our experiments focus on just one intermediate level of compositionality, since we define compounds as dependencies between two words. This choice was based primarily on convenience: we could define compounds as constructions of more than just two words, but the large number of these constructions would make the data partitioning heavier computationally. Focusing on just one level of compositional constructions means that our experiments are not exhaustive in this regard, and we hope to assess other levels of compositionality in future work.

Our goal was to create a benchmark that tests generalisation to novel dependency relations in as comprehensively as possible, not selecting some specific types of dependency relations and leaving out other types. However, memory requirements of the data splitting algorithm do not permit us to use all of the atoms and compounds in the distribution divergence calculations, so we opted to leave out the most frequent and the most infrequent lemmas, and the dependency relations that include them. This means that our set of atoms represents a middle section of the distribution, where the head turns into a tail. Therefore, the controlled dependency relations include lemmas both from the head of the distribution and the tail, although neither of the extremes. We have not been able to assess how this particular choice affects the results.

\section{Acknowledgements}
We thank the anonymous reviewers for their insightful comments and feedback, which significantly improved this paper.
The work was supported by the Academy of Finland grant 337073.
The computational resources were provided by Aalto ScienceIT.

\bibliographystyle{acl_natbib}
\bibliography{custom}

\end{document}